\documentclass[conference]{article}

\usepackage{cite}
\usepackage{spconf}
\usepackage{amsmath,amssymb,amsfonts}
\usepackage{algorithmic}
\usepackage{graphicx}
\usepackage{textcomp}
\usepackage{xcolor}
\usepackage{booktabs}
\usepackage{url}
\def\BibTeX{{\rm B\kern-.05em{\sc i\kern-.025em b}\kern-.08em
    T\kern-.1667em\lower.7ex\hbox{E}\kern-.125emX}}

\begin{document}

\title{
Enhancing Dialogue Annotation with Speaker Characteristics Leveraging a Frozen LLM
}

\name{
Thomas Thebaud$^{\star \dagger}$ \qquad 
Yen-Ju Lu$^{\star}$ \qquad 
Matthew Wiesner$^{\star \dagger}$ \qquad
Peter Viechnicki$^{\dagger}$ \qquad
Najim Dehak$^{\star \dagger}$}
\address{$^{\star}$ Electrical and Computer Engineering Department, Johns Hopkins University, Baltimore, MD, USA \\
$^{\dagger}$ Human Language Technology Center of Excellence, Johns Hopkins University, Baltimore, MD, USA }

\maketitle

\begin{abstract}
In dialogue transcription pipelines, Large Language Models (LLMs) are frequently employed in post-processing to improve grammar, punctuation, and readability. We explore a complementary post-processing step: enriching transcribed dialogues by adding metadata tags for speaker characteristics such as age, gender, and emotion. Some of the tags are global to the entire dialogue, while some are time-variant. Our approach couples frozen audio foundation models, such as Whisper or WavLM, with a frozen LLAMA language model to infer these speaker attributes, without requiring task-specific fine-tuning of either model. Using lightweight, efficient connectors to bridge audio and language representations, we achieve competitive performance on speaker profiling tasks while preserving modularity and speed. Additionally, we demonstrate that a frozen LLAMA model can compare x-vectors directly, achieving an Equal Error Rate of 8.8\% in some scenarios.
\end{abstract}


\section{Introduction}
With the widespread adoption of voice-assisted technologies, automatic transcription services, and real-time speech translation, speech processing tasks have become increasingly prevalent in both consumer and industrial applications.
While automatic speaker recognition plays a crucial role in biometric authentication, forensic analysis, and personalized systems, speech signals also encode a variety of attributes beyond identity—emotional states~\cite{wani2021comprehensive}, demographic information such as age, gender~\cite{hechmi2021voxceleb}, and accent~\cite{deshpande2005accent}.

Recent advancements in deep learning have enabled the use of foundational models such as HuBERT~\cite{hsu2021hubert}, Wav2Vec~\cite{baevski2020wav2vec}, or WavLM~\cite{chen2022wavlm}, which capture rich linguistic and paralinguistic information, as explored and measured in the SUPERB Benchmark~\cite{yang2021superb}.
Despite these successes, typical usage of audio foundation models remains restricted to a narrow range of downstream tasks.

In parallel, large language models (LLMs) like GPT~\cite{openai2024gpt4technicalreport} and LLaMA~\cite{touvron2023llama} have demonstrated exceptional performance across a wide range of text-based tasks—including summarization, information extraction, and dialogue systems—by leveraging vast world knowledge and contextual reasoning capabilities. This has led to a surge in systems that enhance raw audio transcripts with the help of LLMs, improving coherence, disambiguation, and even formatting~\cite{li2024transcription, hu2024large,adedeji2024sound}. However, these systems typically operate only on textual input, without incorporating additional context that can be inferred from the speech signal itself.

Various approaches have been explored to align the representations of audio and text foundation models with the goal of providing LLMs access to audio information other than from transcripts~\cite{gong2023joint, li2024audio, shu2023audio, chu2023qwen, hu2024wavllm}.
The prohibitive cost of fine-tuning all LLM parameters has motivated the use of alternative fine-tuning techniques such as Low-Rank adaptors (LoRA)~\cite{hu2021lora} and variants~\cite{dettmers2024qlora}, which instead train only a small number of new parameters to perform new tasks.
This still implies the fine-tuning of the model for a specific task or set of tasks, which means any addition of a new task will result in either re-starting the fine-tuning process or risking the loss of performance on previous tasks.

To the extent of our knowledge, no prior work has yet explored the use of multimodal LLMs to enrich transcripts with speaker-specific contextual information, such as identity traits, affective state, or other paralinguistic cues—despite the clear utility such information could bring to downstream tasks like diarization, personalization, or inclusive summarization.
In this work, we present a unified framework that bridges audio foundation models with LLMs, while keeping both untouched, enabling us to reuse these pretrained representations for well‐established speech processing tasks, all under a unified, extensible design, with minimal adaptation costs. 

We propose an alternative based on external lightweight connectors that map audio features extracted by a frozen pretrained audio model directly into the embedding space of a frozen LLM, specifically LLAMA-7B fine-tuned via Vicuna~\cite{chiang2023vicuna}. One connector is trained per task, allowing for efficient, modular expansion of capabilities. This architecture enables the LLM to leverage its powerful language reasoning on enriched multimodal inputs, such as audio embeddings encoding speaker characteristics, without retraining or interfering with previous tasks.

In this paper, we make the following contributions:
\begin{itemize}
    \item We propose a modular framework that bridges pretrained audio foundation models with a frozen LLM via external, task-specific, lightweight connectors. This method requires no fine-tuning of either the audio model or the LLM, enabling plug-and-play training of new tasks while preserving performance on existing ones.
    \item We extend this work to the speaker's comparisons within a conversation to lay the groundwork of conversation analysis, by asking an LLM to perform speaker verification tasks, answering questions such as "\textit{Did this speaker speak at least once in the following ten sentences?}".
\end{itemize}

Section~\ref{sec:related_works} surveys related work and highlights key differences with existing approaches, which serve as our baselines. Section~\ref{sec:methods} describes the datasets used, the architecture of our models, and the experimental protocols, including task definitions and evaluation metrics. In Section~\ref{sec:results}, we report the performance of our method across selected tasks, before interpreting the results and discussing broader implications in Section~\ref{sec:conclusion}.

\section{Related Work}
\label{sec:related_works}
In this section, we review prior work related to speech representation learning, the integration of acoustic embeddings into Large Language Models (LLMs), and existing approaches for speaker characterization. We also present the models and techniques that form the foundation of our proposed framework.

\subsection{Accoustic Encoders}
The task of extracting informative and robust representations from raw speech has long been central to speech processing research. Traditional feature extraction methods such as Mel-Frequency Cepstral Coefficients (MFCCs) and spectrograms have been widely used for decades. More recently, self-supervised learning (SSL) approaches have led to significant advances through large-scale transformer-based encoders.

One such model is HuBERT~\cite{hsu2021hubert}, which uses a masked prediction objective to learn powerful latent speech representations. Wav2Vec 2.0~\cite{baevski2020wav2vec} improved upon this by employing contrastive learning, yielding state-of-the-art results in ASR and downstream classification tasks. Building further, WavLM~\cite{chen2022wavlm} introduced a multi-task SSL framework to enhance both supervised and unsupervised performance, particularly excelling in paralinguistic and speaker-centric tasks. The SUPERB benchmark~\cite{yang2021superb} identifies WavLM-large as a leading model for tasks such as speaker identification and emotion recognition.

Additionally, the Whisper model~\cite{radford2023robust}, trained on large-scale multilingual ASR data, has demonstrated cross-domain generalization. Its internal representations have recently been repurposed for tasks such as audio event classification and detection through time and layer-wise Transformer (TLTR) architecture~\cite{gong2023listen}.

\subsection{Audio understanding with LLM}

Large Language Models (LLMs) have increasingly been extended to process non-textual modalities, particularly via fusion with pre-trained audio encoders. These multimodal architectures often rely on parameter-efficient fine-tuning methods such as Low-Rank Adaptation (LoRA)~\cite{hu2021lora}, allowing LLMs to interpret audio-derived embeddings without full model retraining.

Usually fine-tuned using LoRA~\cite{hu2021lora} adaptors, they can be trained to process various encoder embeddings, such as Whisper embeddings for multitasks by LTU-AS~\cite{gong2023joint} or WavLM for emotion recognition~\cite{bellvermultimodal}.
Some recent initiatives also propose models that merge multiple encoders for one or multiple tasks, such as ~\cite{hu2024wavllm}, which leverages both Whisper and WavLM, or even retrain the language model from scratch~\cite{chu2023qwen}.

However, a common limitation in these approaches is their reliance on LLM fine-tuning—either fully or through adaptor layers—which can compromise previous task performances. In contrast, our work investigates a design where both the LLM and the encoder (Whisper or WavLM) remain entirely frozen. Task-specific adaptation is achieved through lightweight external connector modules, preserving the integrity of the base models while enabling extensibility to new speaker-related objectives.

\subsection{Speaker Characterization}
Speaker characterization encompasses a range of tasks aimed at inferring paralinguistic traits from speech, such as identity, gender, age, emotional state, and sociolinguistic attributes. These traits are central to personalization, forensic applications, and conversational analysis.

Recent studies have demonstrated the efficacy of self-supervised models like WavLM in extracting high-dimensional embeddings that capture diverse speech characteristics. For instance, Yang et al.\cite{yang2025demographic} introduced a general classifier based on a fine-tuning of WavLM-large features to infer demographic characteristics, such as age, gender, and native language, from speech. Their framework achieved a Mean Absolute Error (MAE) of 4.94 for age prediction and over 99.81\% accuracy for gender classification across various datasets.

Building upon this, Feng et al.\cite{feng2025vox} proposed Vox-Profile, a comprehensive benchmark designed to characterize rich speaker and speech traits using speech foundation models. Unlike existing work that focused on a single dimension of speaker traits, Vox-Profile provides holistic and multi-dimensional profiles that reflect both static speaker traits (e.g., age, sex, accent) and dynamic speech properties (e.g., emotion, speech flow). The benchmark experiments utilized over 15 publicly available speech datasets and fine-tuned both Whisper large~\cite{radford2023robust} and WavLM large~\cite{chen2022wavlm} for each task.

While these studies have advanced the field of speaker characterization, they primarily rely on fine-tuning SSL models for specific tasks. In contrast, our approach leverages frozen multimodal LLMs and external task-specific adaptors, enabling the integration of speaker traits into downstream applications without the need for retraining the encoder. This modular design facilitates efficient adaptation to new tasks while preserving performance on existing ones.

\section{Methods}
\label{sec:methods}
This section details all the elements used in the experimental pipeline, from the datasets to the metrics used for evaluation.
\subsection{Datasets}
Each task uses a different dataset:

\subsubsection{Automatic Speaker Verification}: 
For the speaker verification task, we use the dev splits of both VoxCeleb1~\cite{nagrani2020voxceleb} and VoxCeleb2~\cite{chung2018voxceleb2}, and we will subsequently evaluate the Equal Error Rate (EER) on the original test split of VoxCeleb1. The VoxCeleb datasets are a collection of celebrity speech segments extracted from YouTube that count respective totals of 1,211 and 5,994 speakers, the VoxCeleb1 test set including 40 independent speakers. They are usually used in automatic speaker verification systems training.

\subsubsection{Speaker Age and Gender Classification}: 
To allow for specific speaker information retrievals, such as age and gender, previous work have proposed automatic labelings of each session, associating each speaker with age and a gender label. For consistency with the baselines~\cite{gong2023joint, yang2025demographic, feng2025vox}, we choose to use the same labels, proposed by~\cite{hechmi2021voxceleb}.

\subsubsection{Speech Emotion Recognition}: 
The IEMOCAP dataset~\cite{busso2008iemocap} is now one of the standard datasets for Speech Emotion Recognition (SER) evaluation, and it contains approximately 12 hours of audio from 10 speakers, annotated with 4 emotions (neutral, happy, angry, sad).

\subsubsection{Automatic Speech Recognition}:
The LibriSpeech~\cite{panayotov2015librispeech} dataset contains 3 training sets of 100, 360, and 500 hours of audio of speakers reading books in a controlled environment. In this study, we will only use the \textit{train-clean-100} split as our train set for the ASR task and the \textit{dev-clean} and \textit{test-clean} as our validation and test sets.
\\

\subsection{Metrics}
\label{sec:metrics}
We evaluate all experiments using a standard set of metrics:
\begin{itemize}
    \item Speech Emotion Recognition (4 classes) and Gender Classification (2 classes\footnote{The dataset considered for evaluation only includes speakers that identify as male or female.}) are evaluated using Accuracy.
    \item Age is evaluated using Mean Average Error (MAE) compared to the ground truth.
    \item Speech Recognition is evaluated using Word Error Rate (WER).
    \item Speaker Verification is evaluated using an Equal Error Rate (EER).
\end{itemize}
For most metrics, we can use the textual outputs of LLAMA, measuring whether the text contains the desired token or the distance with the proposed integer for the age.
For example, for gender classification, 'male' or 'female' are counted as $true$ if they are present for the desired audio; $false$ if they are absent. If both are present in the text output, it is also counted as $false$. 
The exception is the EER, which needs a normalized score to be evaluated.
To measure the EER, we compute the Log Likelihood between the probability of the model using the token 'yes' and the probability of the token 'no' in the output.

\subsection{Models}
In our proposed models, we follow and diverge from the baseline proposed in LTU-AS~\cite{gong2023joint}: 
They used a frozen Whisper model~\cite{radford2023robust}, both to transcribe a given audio into text and to extract the inner representations from the last 32 layers of Whisper.
The inner layers were fed into a TLTR system~\cite{gong2023listen} (pretrained for audio event detection), which applies transformers across the 32 whisper layers and across the time dimension to reduce each audio into one embedding.
The textual prompts, audio embeddings, transcribed text, and expected outputs were then used to fine-tune a pretrained LLAMA Vicuna~\cite{chiang2023vicuna} model using Lora~\cite{hu2021lora}, fine-tuning at the same time the TLTR.

\subsubsection{Speaker Attribute Pipeline}
\begin{figure}[ht]
    \centering
    \includegraphics[width=\linewidth]{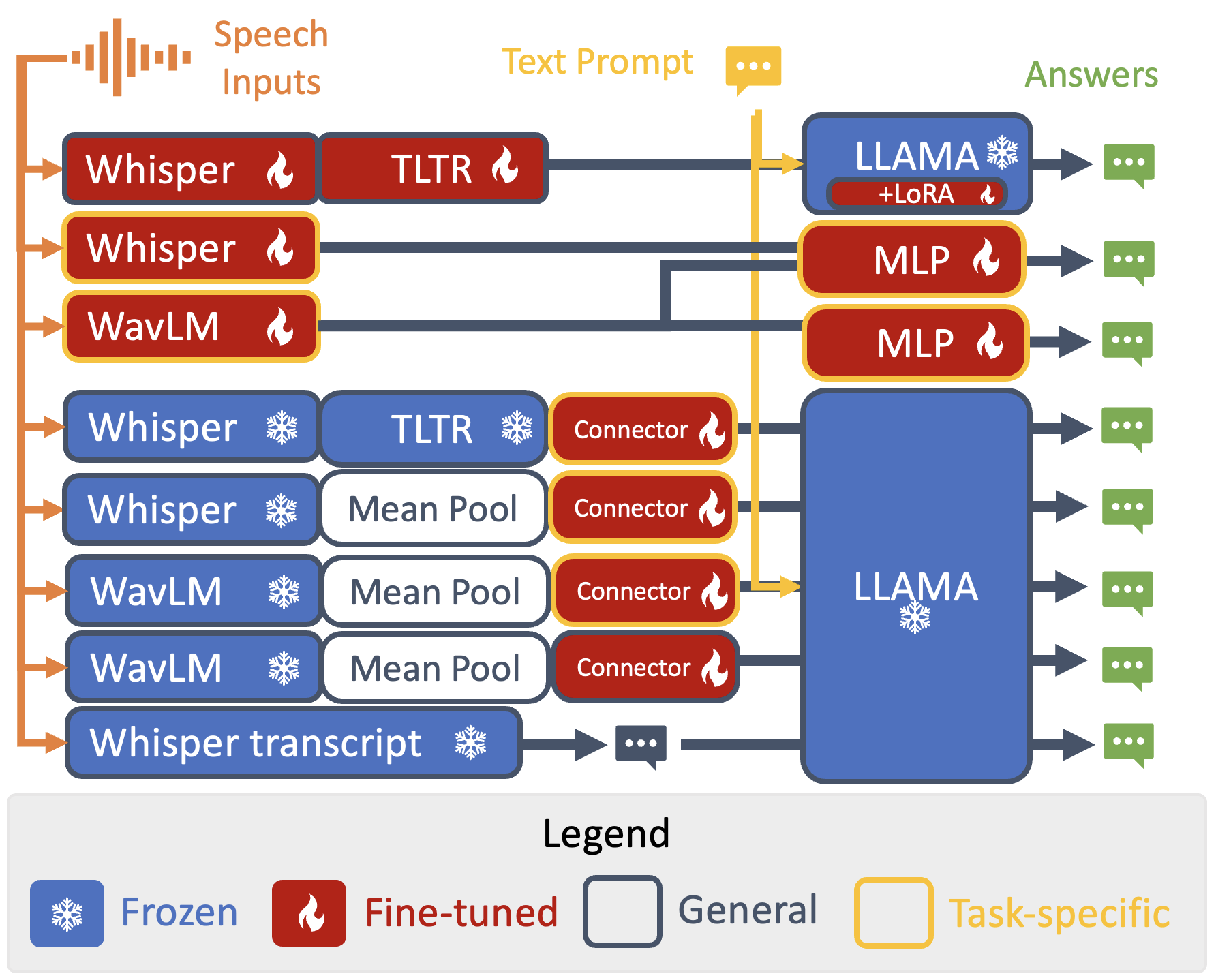}
    \caption{Schematic representing the Speaker Attribute tasks performed. The baseline~\cite{gong2023joint} we are comparing to is the frozen Whisper model with a trainable TLTR connector and a LLAMA fine-tuned using LoRa adaptors.}
    \label{fig:exp1}
\end{figure}

In this article, we focus on the speech settings, so we will consider the performance of the model without using the transcripts, only the audio embeddings.
As shown in Figure \ref{fig:exp1}, we use three audio encoders, frozen: The original TLTR~\cite{gong2023listen}, the last layer of Whisper large v3~\cite{radford2023robust}, and the last layer of WavLM large~\cite{chen2022wavlm}. 
For Whisper and WavLM, we operate mean pooling across the time dimension.
The obtained compressed vectors are projected into the LLAMA's embedding space using a linear connector, before being injected into a pretrained LLAMA Vicuna~\cite{chiang2023vicuna} model.
In all experiments, the only trained parameters are from the linear connectors, one per task.
In both cases, the WavLM encoders remain frozen for the four downstream tasks (ASR, age, gender, and emotion), and only an external connector layer is trained to project the extracted embeddings into LLAMA's embedding space.

\subsubsection{Speaker Verification Pipeline}
As shown in Figure \ref{fig:exp2}, we also perform a speaker verification experiment using a similar framework but with an x-vector encoder.
The architecture of the x-vector is an ECAPA-TDNN~\cite{desplanques2020ecapa}, trained with VoxCeleb1\&2 development sets, using the speechbrain toolkit~\cite{speechbrain} and WavLM representations, which shows an Equal Error Rate(EER) of 0.80\% on the VoxCeleb1-O test set.

\subsection{Experiments}

\subsubsection{Speaker Attributes Extraction}
We begin by evaluating the ability of various pretrained foundational audio models to support speaker attribute inference when paired with a frozen LLAMA language model. The goal is to determine how well speaker traits—such as age, gender, emotion, and linguistic content—can be decoded through this modular framework.

\textbf{Task-specific connectors}:
For each downstream task, we train an independent linear connector that maps the audio encoder’s output into the embedding space of LLAMA. These connectors are trained using prompt-based supervision, where a textual query is paired with the projected audio representation and a target label.
Specifically, the prompts are:
\begin{itemize}
        \item For \textbf{speech emotion recognition}:
    \textit{“What is the emotion of the speaker, using the following audio embeddings: [Audio Embedding Inserted] Emotion: [Label]”}.

    \item For \textbf{age and gender classification}:
    \textit{“What is the age and the gender of the speaker, using the following audio embeddings: [Audio Embedding Inserted] Age: [Label Age] Gender: [Label Gender]”}.

    \item For \textbf{automatic speech recognition} (ASR):
    \textit{“Transcribe the following text: [Audio Embedding Inserted] Transcript: [Label]”}.
\end{itemize}

Model outputs are evaluated on their respective test sets using standard metrics, as defined in Section~\ref{sec:metrics}.

\textbf{Universal connector}:
We compare the performances with those from a universal connector, with the same architecture, but performing all tasks at once, trained with all datasets balanced by a datasampler.

\textbf{Comparison with the transcripts}:
For a better comparison with existing transcription+LLM pipelines, we perform the same set of experiments, but using transcripts extracted with \textit{whisper-large} instead of audio embeddings. 
Whisper shows a 1.4\% EER on Librispeech-test-clean~\cite{radford2023robust}, so the ASR experiment will show how much is lost due to the hallucinations of the version of LLAMA at hand.
Good transcriptions could be revealing of the speaker's gender depending on the linguistic content, and previous work has established that text inputs contain information that helps boost speech emotion recognition tasks~\cite{goncalves2024odyssey}.

\begin{figure}[ht]
    \centering
    \includegraphics[width=\linewidth]{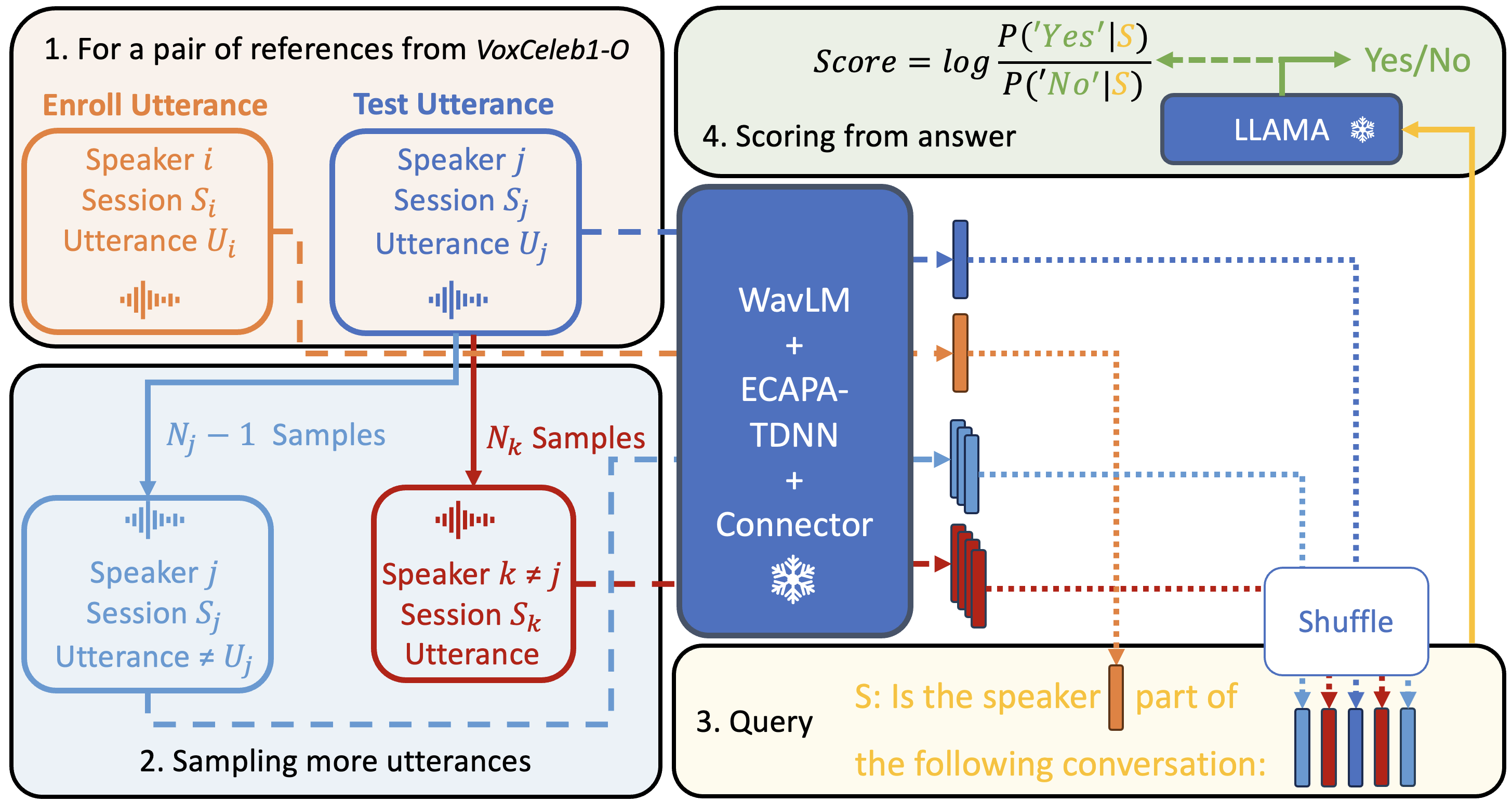}
    \caption{Schematic representing the process to simulate speaker verification within a conversation by taking utterances from the same speaker but from different sessions.}
    \label{fig:exp2}
\end{figure}

\begin{table*}[!ht]
    \centering
    \caption{Performances on various downstream tasks using different encoders, compared to a baseline using only audio inputs. The best value is \textbf{bolded}, and the second best is in \textit{italics}. When different encoders of classifiers are used for different tasks, the number of parameters is shown as $Number Of Systems\times Parameters$\\$\star$VoxProfile does not measure MAE for the age, but accuracy per decade, for which they reached up to 69.5\%.}
    \label{tab:results}
    \begin{tabular}{c l c l c c c c c}
    \toprule
    &                 & Unique &  &  Gender & Age & Emotion & ASR & Trainable \\
    Line & Encoder used    & Encoder & Decoder used & Acc (\%$\uparrow$) & MAE$\downarrow$ & Acc (\%$\uparrow$) & WER (\%$\downarrow$) & parameters \\
    \midrule
    1 & Whisper+TLTR~\cite{gong2023joint} 
                                & $\checkmark$ & LLAMA-7B+LoRA & 95.6  & 8.2   & 58.6  & 97.2 &  49M \\
    2 & WavLM+MLP~\cite{yang2025demographic} 
                                & $\times$ & -  & \textbf{99.81}  & 5.45  & - & - &  $1\times$324M \\
    3 & WavLM+Whisper+MLP~\cite{feng2025vox} & $\times$ & -  &  98.0  & $\star$  & \textit{64.44} & - &  $2\times$1.87B \\
    \midrule
    4 & Whisper+TLTR+Linear \textit{(ours)}     & $\times$ & LLAMA-7B Frozen  & \textit{98.53} & 5.95  & 50.02 & 92.04 & $4\times$33.5M \\ 
    5 & Whisper+Meanpool+Linear \textit{(ours)}   & $\times$ & LLAMA-7B Frozen  & 97.64 & 10.18 & 53.50 & \textit{28.94} & $4\times$10.5M\\ 
    6 & WavLM+Meanpool+Linear \textit{(ours)}    & $\times$ & LLAMA-7B Frozen  & 98.23 & \textbf{2.54}  & \textbf{64.97} & \textbf{27.62} & $4\times$8.3M\\ 
    \midrule
    7 & Whisper Transcription   & $\checkmark$ & LLAMA-7B Frozen & 70.55 & 34.09 & 10.29 & 4.04 & 0 \\
    \bottomrule
    \end{tabular}
\end{table*}

\subsubsection{Speaker Verification}
To push further in the direction of new tasks that could be attributed to LLMs, we explore the automatic speaker verification task by asking the model \textit{Answer by yes or no, are those two audio embeddings from the same speaker:[Audio Embedding Speaker 1] [Audio Embedding Speaker 2] Answer: \{Yes/No\}}.
The audio encoder for this task is a frozen ECAPA-TDNN trained on VoxCeleb1\&2.
The connector is trained using VoxCeleb1 and 2 dev splits to project the ECAPA-TDNN embeddings into LLAMA's embedding space.
For each batch, half of the pairs of embeddings are from the same speaker, and the other pairs are from different speakers.

To evaluate this approach, we compute the answer of LLAMA for each pair of trials from the VoxCeleb1-O test split as a '\textit{yes}' or a '\textit{no}'
However, to compute an EER, we need a score, so we use the priors of LLAMA's outputs to compute the likelihood of the answer, given the input sentence $S$, to have a 'yes' token ($P('yes'|S)$), and the likelihood of it containing a 'no' ($P('no'|S)$).
Then, we compute the log likelihood between both hypotheses and use it as a score. 
This score for each trial allows us to compute the EER as predicted by a Language Model.

To extend this task to additional questions such as '\textit{Was this speaker present in the following conversation?}', we propose an evaluation protocol, using the model trained for Speaker Verification, to ask to compare one audio embedding from a speaker $i$, to a sequence of $N_j\geq1$ and $N_k\geq0$ shuffled embeddings from 2 speakers $j$ and $k$, with $k\neq i$. 
We compute all the embeddings from the same session for a given speaker to simulate a conversation between 2 speakers.
The case $N_j=1 \& N_k=0$ corresponds to the situation evaluated previously. Now, we also evaluate it for various lengths to show how our model behaves in a zero-shot manner when confronted with increasingly difficult settings.
Figure \ref{fig:exp2} illustrates that process.

\section{Results}
\label{sec:results}
This section presents the results obtained first for the speaker attributes experiments, then with the speaker verification experiments.
\subsection{Speaker Attibutes Results}
\subsubsection{Task-specific connectors}
Table~\ref{tab:results} presents the performance of our proposed models across four downstream tasks: gender classification, age prediction, emotion recognition, and automatic speech recognition (ASR). We report results for models based on WavLM and Whisper embeddings, with lightweight task-specific connectors, and compare them against state-of-the-art baselines such as LTU-AS~\cite{gong2023joint}, WavLM+MLP~\cite{yang2025demographic}, and Vox-Profile~\cite{feng2025vox}. The table also includes the number of trainable parameters per model variant.

Overall, our approach demonstrates competitive performance across all speaker-related tasks, with significantly fewer trainable parameters and no fine-tuning of either the acoustic encoder or the language model.

\textbf{Gender Classification:}
The best performance was achieved by our Whisper-based model using a TLTR connector, reaching an accuracy of 98.53\%. This slightly outperformed the WavLM-based model (98.23\%) and surpassed the LTU-AS baseline (95.6\%). Despite its simplicity, our linear connector performs competitively while remaining lightweight.

\textbf{Age Prediction:}  
The WavLM+Meanpool+Linear model outperformed all others, achieving a Mean Absolute Error (MAE) of 2.54 years, significantly better than the Whisper-based variant (5.95 years) and the WavLM+MLP baseline (5.45 years). This suggests WavLM captures more fine-grained speaker age cues than Whisper.

\textbf{Speech Emotion Recognition:}  
WavLM again led performance with 64.97\% accuracy, matching the best baseline (LTU-AS at 58.6\%) and demonstrating the strength of self-supervised embeddings for paralinguistic tasks. In contrast, Whisper-based models underperformed in emotion classification. 

\textbf{Automatic Speech Recognition (ASR):}  
As expected, all models performed poorly on ASR, with Word Error Rates (WER) exceeding 90\%. Whisper performed slightly better than WavLM (28.94\% vs. 27.62\%), though the difference is marginal and likely not statistically significant. Interestingly, part of that high WER is due to the instability of the LLM used, as 23.47\% of the outputted lines were empty for the WavLM experiment, impacting the resulting performances, while none of the outputted lines from the Whisper encoder were empty.

Notably, the Whisper encoder generally underperforms for speaker attribute tasks. This is likely due to its internal pooling mechanism, which emphasizes temporal alignment for transcription but suppresses global paralinguistic cues. In contrast, WavLM embeddings retain broader speaker-dependent characteristics that prove beneficial for age, gender, and emotion tasks.
These results highlight the effectiveness of using frozen acoustic encoders combined with lightweight, task-specific connectors to a frozen LLM. Our approach achieves strong performance while minimizing parameter counts, making it scalable and modular for future applications.

\subsubsection{Universal connector}
We choose not to show the metrics associated with the universal connector, as none of the models were able to learn both the type of expected answer and how to extract them at once, whether using the Whisper+TLTR encoder, the Whisper encoder, or the WavLM encoder.
The common behavior for all models, across the range of hyperparameters explored, was to learn the structure of the answer but provide a constant answer for every query of a certain type. For example, when queried \textit{“What is the age and the gender of the speaker, using the following audio embeddings:"}, the answer was systematically: \textit{"Age: 20 Gender: male”}.
The obvious conclusion is that the complexity of the proposed framework is too simple to see the emergence of a general behavior, and that new directions should be explored, such as more complex and heavy connectors, and allowing for the fine-tuning of other parts of the pipeline.

\subsubsection{Comparison with the transcripts}
The comparative results obtained using the whisper transcripts are shown in Line 7 of Table \ref{tab:results}.
The ASR task shows a non-negligeable degradation of the WER, from 1.4\% in the whisper transcripts relative to the ground truth, to 4.04\% on the outputs of the LLAMA Vicunas. 
The slight degradation observed does not explain entirely the high WER observed using the WavLM and Whisper embeddings mean pooled, which supports the idea of using parallel transcripts in addition to audio embeddings for future dialogue annotations.
As expected, age, gender and emotion predictions are significantly worse than previous experiments, with notably emotions being significantly worse than random (10.29\% accuracy for 4 classes) and gender being almost random, knowing that 73\% of the utterances in the set are Male. The F1-score for gender classification is measured 0.08.

\subsection{Speaker Verification Results}

\begin{table}[!ht]
    \centering
    \caption{Performances on the speaker verification task, considering various numbers of embeddings for each test speaker. We always use one enrollment embedding from a speaker $i$ to compare to $N_j+N_k$ test embeddings. $N_j\geq1$ is the number of embeddings used for the target speaker, while $N_k\geq0$ is the number of embeddings used for a third speaker $k$, different than both $i$ and $j$.}
    \label{tab:asv_results}
    \begin{tabular}{c c c c}
        \toprule
         Line & $N_j$ & $N_k$ & EER$\downarrow$ (\%) \\
         \midrule
         1 & 1 & 0 & 12.08 \\
         2 & 5 & 0 & 11.44 \\
         3 & 1 & 5 & 10.34 \\
         4 & 5 & 5 & 8.80 \\
         \bottomrule
    \end{tabular}
\end{table}


Table \ref{tab:asv_results} presents our speaker verification experiments results.
Line 1 shows the standard one-to-one comparison used for speaker verification, following the couple of enroll/test audio segments of VoxCeleb1-O.
On this task, our adapted system yields 12.08\% EER, a very high result compared to the original system that yielded 0.80\% on the same setup. 
This first result shows two conclusions: 
\begin{enumerate}
    \item Without fine-tuning nor cosine products, a LLAMA model cannot reproduce as precisely the comparisons between speakers, and by far.
    \item However, the comparison is possible, as 12.08\% is far from random. By comparison, pre-neural systems such as GMM-UBM~\cite{liu2006improved} and $i$-vectors~\cite{dehak2010front} show respectively 15.0\% EER and 8.8\% EER~\cite{nagrani2020voxceleb} when trained on VoxCeleb1-dev and evaluated on VoxCeleb1-O.
\end{enumerate}

Then, comparing line 1 with line 2 and line 3 with line 4, by adding more content about the targeted speaker (always from the same session), the model becomes more precise, which is expected.
When adding the presence of a different speaker, the performance also improves (comparing lines 1-2 with lines 3-4).
This is explained by the concept behind the EER: being defined as the value where the False Acceptance Rate and False Rejection Rates equal, making the rejection easier by adding more non-target embeddings to the negative cases lowers the False Rejection Rate, which incidentally lowers the EER.

\section{Conclusion}
\label{sec:conclusion}

In this work, we presented a modular and scalable framework for speaker-centric dialogue annotation by bridging frozen audio foundation models with a frozen large language model (LLAMA-7B Vicuna) using lightweight, task-specific connectors. Our approach avoids any fine-tuning of the base models, relying solely on simple linear adaptors to project audio-derived embeddings into the LLM's latent space.

We evaluated this architecture across four representative tasks: gender classification, age prediction, emotion recognition, and automatic speech recognition. Despite using only a fraction of the trainable parameters compared to LoRA-based or fully fine-tuned systems, our method achieved competitive performance on all attribute-based tasks. These results demonstrate that pretrained LLMs, when augmented with minimal trainable components, can successfully perform speech processing tasks from audio-derived representations.

Additionally, we explored a novel formulation of the speaker verification task using LLMs, enabling 1-to-N comparisons of speaker embeddings within simulated multi-turn conversations. While the EER of 12.08\% for one-to-one verification is substantially higher than the fine-tuned ECAPA-TDNN baseline (0.80\%), it is nonetheless significantly better than random and aligns with earlier pre-neural systems. 
Notably, verification accuracy improved as more in-domain context (e.g., multiple utterances or distractor embeddings) was provided, suggesting that LLMs possess latent potential for conversational speaker modeling under richer contexts.

Our method offers a lightweight, adaptable alternative to current multimodal pipelines, enabling rapid extension to new speaker-related tasks without compromising previously established language processing performances. However, its reliance on highly specialized, task-specific connectors limits generalization: attempts to unify multiple tasks under a simple shared connector architecture failed to converge, highlighting the need for more complex connector designs.

In future work, we plan to explore the integration of more powerful and flexible connector modules, such as X-Formers~\cite{swetha2024x} or Q-Formers~\cite{liu2024visual}, as well as to incorporate prompt tuning or lightweight finetuning of the LLM itself. 
We are also particularly interested in leveraging the newly released LLAMA 3.3 series models\footnote{\url{https://www.llama.com/}}, which offer larger capacity and stronger alignment, as a foundation for more advanced conversational understanding.

\bibliographystyle{IEEEtran}
\clearpage
\bibliography{main}

\end{document}